\documentclass[runningheads]{llncs}
\usepackage[T1]{fontenc}
\usepackage{graphicx}
\usepackage{textcomp}
\usepackage{xcolor}
\def\BibTeX{{\rm B\kern-.05em{\sc i\kern-.025em b}\kern-.08em
    T\kern-.1667em\lower.7ex\hbox{E}\kern-.125emX}}

\usepackage{multirow}
\usepackage{subcaption}
\usepackage{amsmath,amsfonts,amssymb}
\usepackage{algorithm}%
\usepackage{tikz}
\usetikzlibrary{automata,positioning}
\usepackage{forest}
\usepackage{caption}
\usepackage{subcaption}
\usepackage{hyperref}
\usepackage{rotating}

\usepackage{lscape}
\usepackage{rotating}
\usepackage{multicol}

\usepackage{array}
\usepackage{orcidlink}

 \mathchardef\mhyphen="2D

\begin{document}
\title{Discriminative Rule Learning for Outcome-Guided Process Model Discovery}
%
%
\author{Ali Norouzifar \orcidlink{0000-0002-1929-9992} \and
Wil van der Aalst \orcidlink{0000-0002-0955-6940}}
\authorrunning{A. Norouzifar et al.}
%
\institute{RWTH University, Aachen, Germany \\ \email{\{ali.norouzifar, wvdaalst\}@pads.rwth-aachen.de}}

%
\maketitle              
\begin{abstract}
Event logs extracted from information systems offer a rich foundation for understanding and improving business processes. In many real-world applications, it is possible to distinguish between desirable and undesirable process executions, where desirable traces reflect efficient or compliant behavior, and undesirable ones may involve inefficiencies, rule violations, delays, or resource waste. This distinction presents an opportunity to guide process discovery in a more outcome-aware manner. Discovering a single process model without considering outcomes can yield representations poorly suited for conformance checking and performance analysis, as they fail to capture critical behavioral differences. Moreover, prioritizing one behavior over the other may obscure structural distinctions vital for understanding process outcomes.
By learning interpretable discriminative rules over control-flow features, we group traces with similar desirability profiles and apply process discovery separately within each group. This results in focused and interpretable models that reveal the drivers of both desirable and undesirable executions. The approach is implemented as a publicly available tool and it is evaluated on multiple real-life event logs, demonstrating its effectiveness in isolating and visualizing critical process patterns. 
\end{abstract}
\section{Introduction}
Event logs extracted from information systems provide a valuable foundation for analyzing and improving business processes. Process discovery aims to automatically derive process models from event logs, enabling further analyses such as conformance checking and performance optimization. In many real-world applications, process executions can be distinguished as desirable or undesirable. While desirable traces reflect efficient and compliant behavior, undesirable ones may involve inefficiencies, violations, delays, or resource waste. This distinction presents an opportunity: desirable traces illustrate desirable behavior, whereas undesirable ones reveal what should be avoided.

A common strategy in such contexts is to discover a process model that supports desirable traces while excluding undesirable ones~\cite{DBLP:conf/sac/NorouzifarA23}. However, a single global model often obscures the differences between the desirable and undesirable behavior of the processes. It lacks the capacity to reveal how different execution paths contribute to process outcomes or what specific control-flow characteristics distinguish desirable behavior from undesirable behavior.

In this paper, we propose a method to bridge this gap by identifying key control-flow patterns that effectively distinguish between desirable and undesirable traces. To this end, we introduce a methodology for selecting a subset of strong discriminative patterns, ensuring that only the most representative pattern is retained among those satisfied by similar trace subsets. Based on the selected patterns, we group traces and apply process discovery independently within each group. This leads to focused and interpretable process models that more accurately reflect the behavioral characteristics of distinct trace subsets. Our approach not only highlights the control-flow structures associated with trace desirability but also uncovers the underlying drivers of both desirable and undesirable process behaviors.

\begin{figure}[t]
    \centering
    \begin{minipage}[t]{0.48\textwidth}
    \vspace{-60pt}
        \centering
         \captionof{table}{\small Trace variants and frequencies in \( L^+ \) and \( L^- \).}
        \label{tab:trace_variants}
        \begin{tabular}{|c|c|c|}
            \hline
            \textbf{Trace Variant} & \textbf{Freq in \( L^+ \)} & \textbf{Freq in \( L^- \)} \\
            \hline
            $\langle p,a,l \rangle$ & 100 & 0 \\
            $\langle p,l,a \rangle$ & 100 & 0 \\
            $\langle a,p,l \rangle$ & 50  & 50 \\
            $\langle a,l,p \rangle$ & 50  & 50 \\
            $\langle l,a,p \rangle$ & 0   & 100 \\
            $\langle l,p,a \rangle$ & 0   & 100 \\
            \hline
        \end{tabular}
    \end{minipage}%
    \hfill
    \begin{minipage}[t]{0.48\textwidth}
        \centering
        \includegraphics[width=\linewidth]{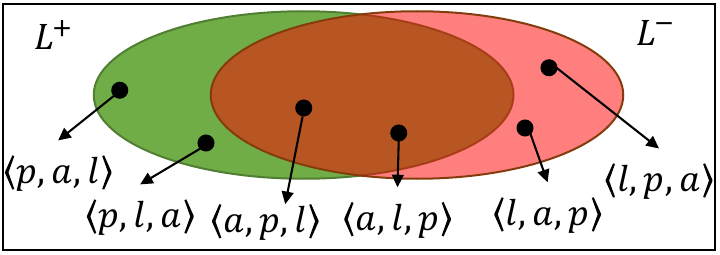}
        \caption{\small Distribution of trace variants across desirable \(L^+\) and undesirable \(L^- \) event logs.}
        \label{fig:Lp_Lm_van}
    \end{minipage}
    \vspace{-20pt}
\end{figure}

Consider an order-handling process that involves three distinct sequential activities: \textbf{($p$)} \textbf{p}ick items from shelves. \textbf{($a$)} \textbf{a}ssemble package by preparing for shipment. \textbf{($l$)} Print shipping \textbf{l}abel.
Ideally, orders follow a logical workflow such as \(\langle p, a, l \rangle\) or \(\langle p, l, a \rangle\). 
An analysis of historical data reveals two distinct sets of cases: a desirable event log \(L^+\) representing workflows with an on-time delivery, and an undesirable event log \(L^-\) capturing inefficient or problematic workflows. The frequency of trace variants in each event log is shown in Table~\ref{tab:trace_variants}. Figure~\ref{fig:Lp_Lm_van} positions the traces based on their belonging to desirable or undesirable behavior of the process.

\begin{figure}[tb]
\centering
\begin{minipage}[t]{0.53\linewidth}
    \centering
    \begin{subfigure}[t]{0.6\linewidth}
        \centering
        \includegraphics[width=\linewidth]{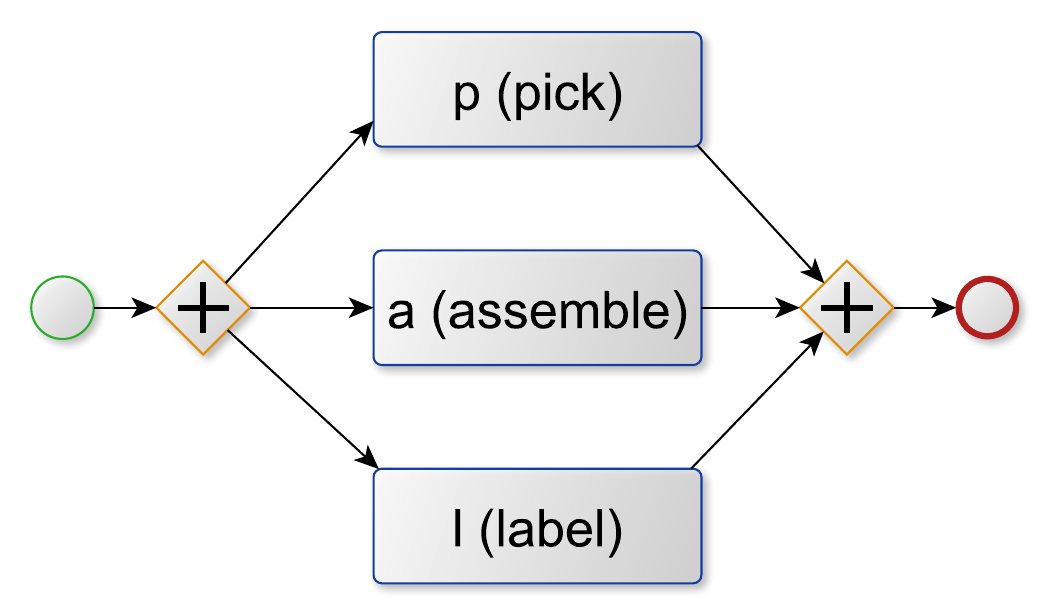}
        \caption{\small Discovered process model for \( L^+ \) or \( L^+ \cup L^-\).}
        \label{fig:imf_model}
    \end{subfigure}
\end{minipage}%
\hfill
\begin{minipage}[t]{0.45\linewidth}
    \vspace*{-80pt} 
    \begin{subfigure}[t]{\linewidth}
        \centering
        \includegraphics[width=0.8\linewidth]{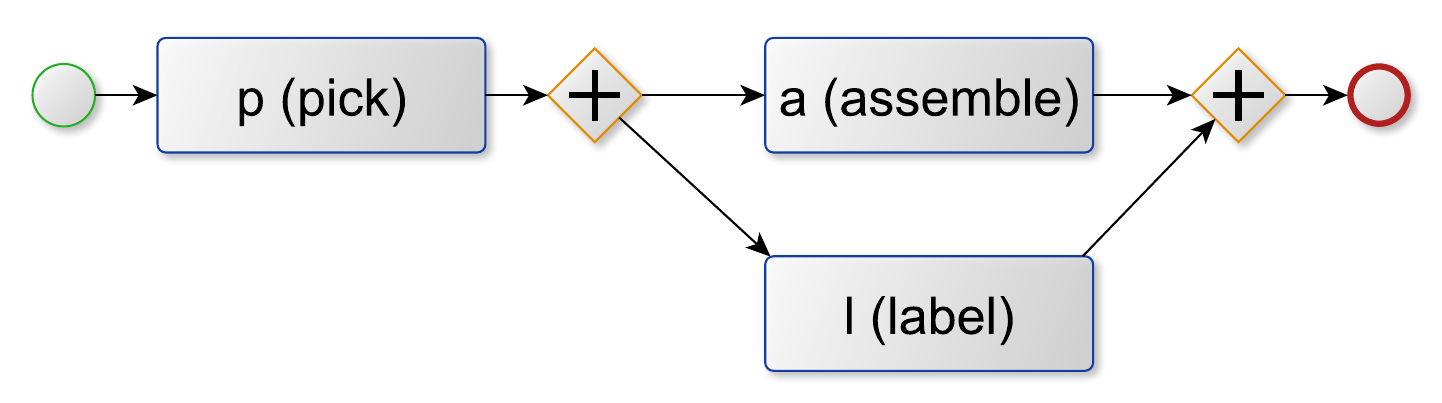}
        \caption{\small Discovered model for \( L^+ \).}
        \label{fig:motiv_des_model}
    \end{subfigure}
    
    \vspace{1em} 
    \begin{subfigure}[t]{\linewidth}
        \centering
        \includegraphics[width=0.8\linewidth]{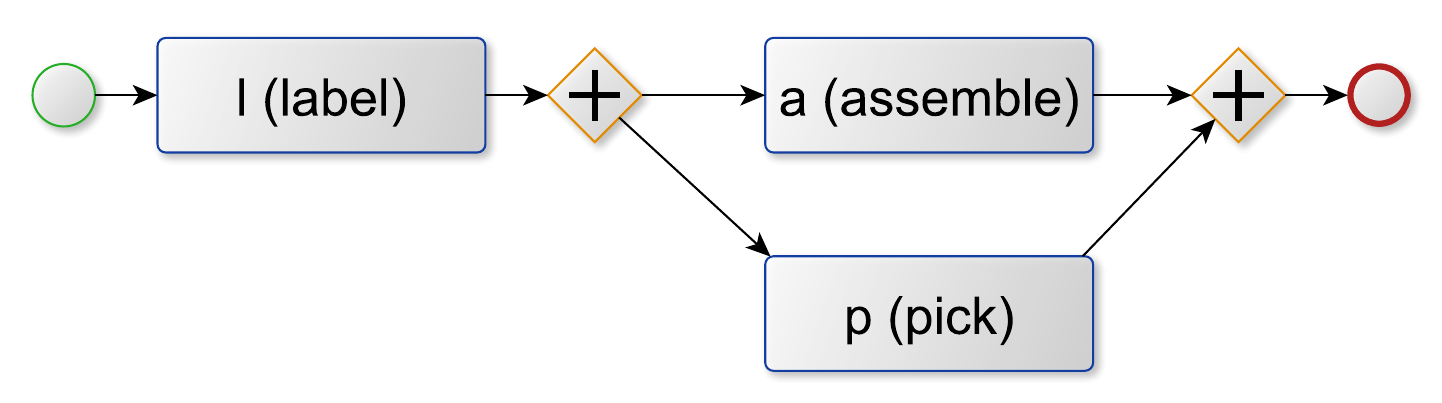}
        \caption{\small Discovered model for \( L^- \).}
        \label{fig:motiv_undes_model}
    \end{subfigure}
\end{minipage}
\caption{\small  Process models for desirable and undesirable traces.}
\label{fig:motivating_models}
\vspace{-10pt}
\end{figure}

Although \(L^+\) and \(L^-\) share overlapping trace variants, their differing frequency distributions reflect distinct process behaviors. Discovering a process model that accounts for all observed traces without considering trace desirability can result in the model shown in Fig.~\ref{fig:imf_model}. Even if the model is discovered using only \(L^+\), it may produce the same generalized process model, especially if sequences such as \(\langle l, a, p \rangle\) and \(\langle l, p, a \rangle\) are considered plausible but unobserved (i.e., allowed by generalization). As a result, the discovered model permits all possible orderings of the activities and achieves perfect fitness with respect to both the desirable and undesirable event logs, while failing to capture their behavioral differences.

\begin{figure}[tb]
    \centering
    \includegraphics[width=0.65\linewidth]{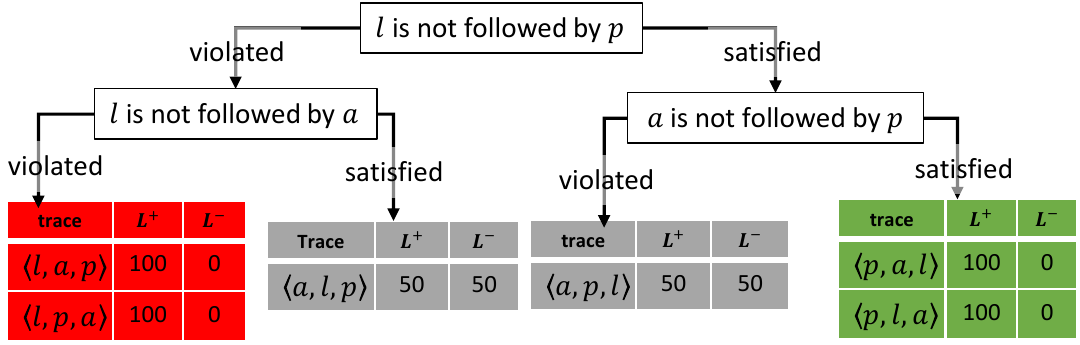}
    \caption{\small Decision tree classifying traces into desirable or undesirable.}
    \label{fig:dt_model}
    \vspace{-25pt}
\end{figure}

We leverage the predictive power of supervised learning models to identify discriminative control-flow characteristics that distinguish desirable from undesirable traces. Traces are first encoded into a feature space constructed from declarative constraints, which effectively capture relationships between activities. The most informative features are then selected to explain behavioral differences, and subsequently used to extract and filter discriminative process variants that represent the core distinctions between the two classes.

A simple decision tree, shown in Fig.~\ref{fig:dt_model}, suggests that traces are likely to be classified as desirable if \(l\) is not eventually followed by \(p\), and \(a\) is also not eventually followed by \(p\). Based on this logic, the process model depicted in Fig.~\ref{fig:motiv_des_model} captures the control-flow of a subset of desirable traces, covering 66.7\% of \(L^+\) while excluding all traces from \(L^-\). 
Unlike the generalized model in Fig.~\ref{fig:imf_model}, which permits all observed behavior including undesirable variants, this model emphasizes the representation of preferred behavior while explicitly avoiding undesirable patterns.

Conversely, Fig.~\ref{fig:motiv_undes_model} presents a process model focused on undesirable traces, corresponding to the red-labeled subgroup of traces in Fig.~\ref{fig:dt_model}. This model covers 66.7\% of \(L^-\) while excluding all traces from \(L^+\), effectively capturing behavior specific to undesirable executions. This example demonstrates how discriminative control-flow features can guide the extraction of focused process variants that clearly separate desirable and undesirable behavior. While the other branches of the decision tree in Fig.~\ref{fig:dt_model} could also be used to derive additional process models, their associated subsets do not exhibit a dominant desirability class.

In real-life scenarios, effectively guiding process discovery requires a comprehensive approach that includes selecting a meaningful feature space to capture activity relationships, choosing a supervised learning model that balances predictive power with interpretability, and leveraging the model’s explanations to derive event logs focused on the distinguishing aspects. These targeted logs can then be used to discover process models that explain the underlying behavioral differences. This paper presents such an approach in detail.

\vspace{-5 pt}
\section{Related Work} 
Declarative process discovery focuses on identifying a set of constraints that restrict the allowed behavior, ranging from no constraints permitting all behavior to highly restrictive combinations that allow no behavior at all. Some approaches, such as~\cite{DBLP:journals/tkde/ChesaniFGLMMMT23} and~\cite{DBLP:conf/edoc/ChesaniFGGLMMMT22}, begin with a model that permits only desirable traces and iteratively introduce declarative constraints to exclude undesirable behavior. Another example is the rejection miner proposed in~\cite{DBLP:journals/is/SlaatsDBC24}, which selects patterns from a given input set that are satisfied by desirable traces and then minimizes this set to include patterns that reject undesirable traces. While such methods effectively yield declarative models that characterize forbidden behavior in the process, our approach differs in focus: we aim to identify discriminative patterns and subsequently discover imperative models that describe how process executions are actually carried out.

The discovery of imperative process models, such as BPMNs or Petri nets, represents another important line of research. For example, the approach introduced in~\cite{DBLP:journals/isci/LeonNCB18} and~\cite{DBLP:journals/jmlr/GoedertierMVB09} do not assume explicit access to undesirable traces. Instead, they generate artificial negative events to approximate undesired behavior and use these to evaluate generalization and guide the discovery process away from such behavior. A different line of work, such as~\cite{DBLP:conf/sac/NorouzifarA23}, incorporates both desirable and undesirable event logs in a recursive process discovery framework, where each iteration seeks a process structure that effectively captures the desirable behavior while avoiding the undesirable one.

Clustering techniques without considering the desirability of cases can help to reduce the complexity of the control-flow by grouping similar behaviors. However, they do not guarantee that desirable and undesirable traces are assigned to separate clusters~\cite{DBLP:journals/tkde/WeerdtBVB13}. Other approaches, such as process variant identification~\cite{DBLP:conf/bpmds/NorouzifarRDA24} and concept drift detection~\cite{DBLP:journals/csur/SatoFBS22}, aim to identify control-flow variability across different dimensions, such as the time or performance dimensions.

Comparative process analysis offers a variety of techniques to analyze and contrast different subprocesses~\cite{DBLP:journals/is/BoltLA18}. 
The field of predictive process monitoring investigates supervised learning approaches capable of effectively classifying different groups of cases~\cite{DBLP:journals/tkdd/TeinemaaDRM19}. While these techniques are highly effective for prediction tasks, they often lack interpretability and are not intended to serve as descriptive models of process behavior.
The area of deviance mining explores the application of supervised learning methods to identify the key factors that distinguish normal from deviating cases~\cite{BISE_dev_mining}. These approaches aim to uncover the driving features behind behavioral deviations, providing valuable insights into the root causes of anomalies. 

The deviance mining pipeline introduced in~\cite{BISE_dev_mining} constructs a feature space by combining declarative constraints, sequential features, and data attributes to represent traces. White-box supervised models, such as decision trees, are then trained to classify traces, and the paths from root to leaf nodes are reported as the most relevant features contributing to the classification. In contrast, our approach treats such rules as an intermediate feature space that captures relationships among features. On top of this space, we train a sparse regression model capable of assigning importance scores to the rules, thereby identifying those that contribute most to the classification and can serve as explanations of the differences between variants. We further cluster the rules and leverage the most distinguishing ones to project event logs and discover process models that represent the variants satisfying the corresponding distinguishing rules.

In~\cite{DBLP:conf/icpm/RafieiBPPHLA23}, a decision tree-based rule extraction approach is introduced to support study planning. However, the robustness and generalization ability of single decision trees for identifying discriminative rules can be limited. In this paper, we propose an enhanced rule extraction method based on ensemble tree models, combined with importance scoring derived from a supervised linear regression model. Our approach is inspired by the framework presented in~\cite{DBLP:conf/icml/Cohen95}, and leverages declarative feature extraction to capture control-flow characteristics. These insights are subsequently translated into interpretable process models using imperative process discovery techniques.

\vspace{-10pt}
\section{Preliminary}
Given a set of elements \(A\), the notation \(\mathcal{P}(A)\) refers to the power set of \(A\), $A^*$ refers to all sequences we can generate over the elements of this set and $A^n$ with $n \in \mathbb{N}$ is the universe of all vectors with $n$ elements from set $A$. Considering a vector $b\in A^n$, $b[i]$ refers to the $i$-th element of the vector where $1 \leq i \leq n$.
In the following, we introduce an event log formally.

\begin{definition}[Event Log]
Let \(\mathcal{C}\) denote the universe of case identifiers and \(\mathcal{A}\)  denote the universe of activities. 
A trace is a tuple $\sigma=(c,t)$, where $\pi_c(\sigma)=c \in \mathcal{C}$ is the case identifier and $\pi_{t}(\sigma)=t \in \mathcal{A}^*$ is the control flow of this trace. 
An event log $L \subseteq \mathcal{C} \times \mathcal{A}^*$ is a set of traces such that the case identifier of each trace is unique, i.e., considering any $\sigma, \sigma^{\prime} \in L$, $\pi_c(\sigma) \neq \pi_c(\sigma^{\prime})$. $\mathcal{L}$ is the universe of all event logs.
\end{definition}

For instance, 
\(L{=}\{(1,\langle p,a,l \rangle), (2,\langle a,l,p \rangle), (3,\langle p,a,l \rangle), (4,\langle p,a,l \rangle), (5,\langle a,l,p \rangle), \) \((6,\langle l,p,a \rangle) \} \) is an event log.

\begin{definition}[Case Label]
Let \(L \in \mathcal{L}\) be an event log. \(\kappa_L: L \rightarrow \{0, 1\}\) is a trace labeling function that assigns 0 to a trace if it is undesirable and 1 if it is desirable. This function is assumed to be given as input. If the context is clear, we omit the subscript \(L\) and write \(\kappa(\sigma)\) for $\sigma \in L$. 
\end{definition}

For example, 
\(\kappa {=} \{((1,\langle p,a,l \rangle),1), ((2,\langle a,l,p \rangle),0), ((3,\langle p,a,l \rangle),1), ((4,\langle p,a,l \rangle),\) \\ \(1), ((5,\langle a,l,p \rangle),1), ((6,\langle l,p,a \rangle),0) \} \)
is a labeling function with four desirable traces and two undesirable traces.

Declarative constraints are specific instances of predefined constraint templates that are applied to activities extracted from an event log. These constraints capture behavioral relationships and conditions in a flexible and rule-based manner.
There exists a variety of well-established declarative constraint templates, including the following examples:
 $CoExistence(x_1, x_2)$: If activity $x_1$ occurs in a trace, then activity $x_2$ must also occur, and vice versa.
  $AtLeast1(x_1)$: Activity $x_1$ must occur at least once in the trace. 
 $ChainResponse(x_1, x_2)$: If activity $x_1$ occurs, activity $x_2$ must occur immediately afterward.
For a detailed explanation of various declarative constraint templates, refer to~\cite{DBLP:journals/tmis/CiccioM15}.

\begin{definition}[Declarative Constraints]
\label{Def:rules}
Let $\mathcal{T}$ be a non-empty set of templates, where each template is a relation $d(x_1, \dots,$ $x_m) \in \mathcal{T}$ defined over variables $x_1, \dots, x_m$, with $m \in \mathbb{N}$ representing the arity of $d$. For a given set of activities $a_1, \dots, a_m \in \mathcal{A}$, a declarative constraint $d(a_1, \dots, a_m)$ is derived as a specific instantiation of the template $d$, binding its variables to the corresponding activities. $\mathcal{T}_A$ is the universe of declarative constraints that we can define over $A \subseteq \mathcal{A}$.
\end{definition}

Each trace can either satisfy a constraint, violate it, or lack sufficient information to evaluate it (vacuous satisfaction). 

\begin{definition}[Declarative Constraints Evaluation]
Let \(A \subseteq \mathcal{A}\) be a set of activities. Each constraint \(d \in \mathcal{T}_A\) consists of an \emph{activation condition} \(d_{\triangleright}\) and a \emph{target condition} \(d_{\square}\). The activation condition determines when the constraint becomes relevant (i.e., triggered), and the target condition defines the behavior expected once the constraint is activated. Given a trace \(\sigma \in \mathcal{A}^*\), the evaluation of constraint \(d\) over \(\sigma\) yields one of the following three outcomes:
\begin{itemize}
    \item Violation: \(\sigma \nvDash d\) if \(d_{\triangleright}\) is satisfied in \(\sigma\) but \(d_{\square}\) is not.
    \item Satisfaction: \(\sigma \vDash d\) if both \(d_{\triangleright}\) and \(d_{\square}\) are satisfied in \(\sigma\).
    \item Vacuous Satisfaction: \(\sigma \Vert d\) if \(d_{\triangleright}\) is not satisfied in \(\sigma\).
\end{itemize}

Considering {\small \(
\mathcal{S} = \{satisfied,\, violated,\, vac\mhyphen satisfied\}
\)} as the set of possible evaluation outcomes, the evaluation function 
{\small \(
eval_L: L \times \mathcal{T}_{act(L)} \rightarrow \mathcal{S}
\)}
is defined by
{\small
\[
eval_L(\sigma,d) =
\begin{cases}
violated, & \text{if } \sigma \nvDash d,\\[1mm]
satisfied, & \text{if } \sigma \vDash d,\\[1mm]
vac\mhyphen satisfied, & \text{if } \sigma \Vert d.
\end{cases}
\]}
\end{definition}

For example, considering the trace $\sigma$ with $\pi_t(\sigma) = \langle p, l \rangle$, the constraint $CoExistence(a, p)$ is violated, since $a$ does not occur, while $p$ does. The constraint $ChainResonse(a, p)$ cannot be evaluated, as activity $a$ does not appear in the trace (vacuous satisfaction). The constraint $AtLeast1(p)$ is satisfied.

\section{Discriminative Process Variants Discovery}

\begin{figure} [tb]
    \centering
    \includegraphics[width=1\linewidth]{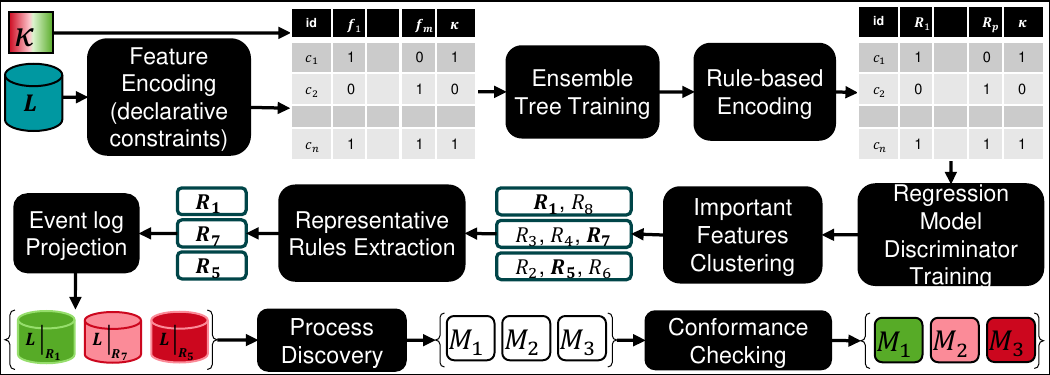}
    \caption{\small An overview of our proposed framework.}
    \label{fig:framework_overview}
    \vspace{-15pt}
\end{figure}

Figure~\ref{fig:framework_overview} illustrates the proposed framework. The process starts by encoding traces with declarative constraints, which provide interpretable control-flow features. Ensemble tree-based models are then trained to capture complex interactions among these features. From the trained models, decision rules are extracted to construct an enriched feature space. A sparse logistic regression model is subsequently trained, and the most important features are clustered based on their similarity in representing traces within the feature space. For each cluster, the most representative rule is selected and used to filter traces, enabling the discovery of focused process variant models. Finally, these models are evaluated using dedicated metrics to ensure they effectively highlight the behavioral contrasts between \(L^+\) and \(L^-\).

\subsection{Feature Encoding}
The algorithm begins by discovering all declarative constraints that are satisfied by at least one trace in the log. These constraints are then pruned using the subsumption hierarchy: when multiple constraints have a subsumption relationship and yield identical evaluations across the event log, we retain only the most restrictive constraint~\cite{DBLP:journals/tmis/CiccioM15}. Consider the function $declare {:} \mathcal{L} {\rightarrow} \mathcal{T_\mathcal{A}}$ as a function extracting such constraints. The resulting set of constraints is used to encode the event log into a structured feature space.

\begin{definition}[Feature Space]
Let $L {\in} \mathcal{L}$ be an event log and {\small $declare(L) \subseteq \mathcal{T}_{act(L)}$} be a set of declarative constraints extracted from it.
    We define the feature space as 
{\small
\(
\mathcal{F}_L = declare(L) \times \mathcal{S}.
\)}
An encoding function 
{\small\(
encode_L: L \rightarrow \{0,1\}^{|\mathcal{F}_L|}
\)}
maps each trace {\small $\sigma \in L$} to a binary feature vector. For every feature {\small $f_i = (d,s) \in \mathcal{F}_L$} where {\small $1 \leq i \leq \vert \mathcal{F}_L\vert$}, the corresponding encoding is defined as
{\small
\[
encode_L(\sigma)[i] =
\begin{cases}
1, & \text{if } eval(\sigma,d)= s,\\[1mm]
0, & \text{otherwise}.
\end{cases}
\]}
\end{definition}

\subsection{Ensemble Tree-Based Feature Extraction}
To enhance the feature space, we further leverage tree ensemble models such as random forests and gradient boosting. In a \emph{random forest}, each decision tree is trained on a bootstrap sample of $L$, and at each node, a random subset of features is considered for splitting. This introduces randomness that reduces variance and improves generalization. \emph{Gradient boosting} constructs trees sequentially, where each tree is trained to correct the residual errors of the ensemble constructed so far, thereby reducing bias.

\begin{definition}[Rule Extraction]
A rule \( R \in \mathcal{P}(\mathcal{F}_L) \) is defined as a set of features. 
Let $\Psi$ denote an ensemble of decision trees, which may be obtained from either a random forest or a gradient boosting model. For each decision tree $\psi \in \Psi$, $\mathcal{R}_{\psi}$ denotes the set of all rules extracted from $\psi$, where every path from the root to a leaf node in a decision tree corresponds to a rule.  
\(
\mathcal{R}_{\Psi} = \bigcup_{\psi \in \Psi} \mathcal{R}_{\psi}
\)
is the set of all rules extracted from an ensemble tree. 
\end{definition}

Considering the decision tree represented in Fig.~\ref{fig:dt_model} as a tree from and ensemble of trees, 
\(
 \{(NotSuccession(l,p),violated), (NotSuccession(l,a),violated)\}
\) is one of the four rules we can extract from this decision tree.

\begin{definition}[Rule-based Encoding]
Let $\Psi$ be an ensemble tree and $\mathcal{R}_{\Psi}$ be the set of all rules extracted from it. The function {\small $\phi_L {:} L {\rightarrow} \{0,1\}^{\vert \mathcal{R}_{\Psi} \vert}$} is a mapping from the traces to the rule-based encoding such that considering {\small $R_i {\in} \mathcal{R}_{\Psi}$},
{\small
\(
\phi_L(\sigma)[i] {=} \underset{f_j\in R_i}{\bigwedge} encode(\sigma)[j]
\)}
, where {\small $1 {\leq} i {\leq} \vert \mathcal{R}_{\Psi} \vert$} and  {\small $1 {\leq} j {\leq} \vert \mathcal{F}_L\vert$}.
    
\end{definition}

\vspace{-10 pt}
\subsection{Regression Model}
This new feature space encapsulates the ensemble-derived features and is used to augment our original feature set. Our methodology combines the strengths of declarative constraint encoding, ensemble tree-based feature extraction, and sparsity-inducing logistic regression to produce interpretable and efficient process models that distinguish between desirable and undesirable behaviors.
\begin{definition}[Regression Classifier]
Let $L \in \mathcal{L}$ be an event log and $L^{\prime} \subseteq L$ be a subset of traces selected for training of the regression model. Considering $\sigma \in L^{\prime}$ as a trace from this event log, a logistic regression classifier with L1 regularization is defined as:
{\small
\[
cls(\sigma) = 
\begin{cases}
1, & \text{if }  g\left(w^\top \phi(\sigma) + b\right) \geq 0.5, \\
0, & \text{otherwise},
\end{cases}
\]}
where \(w \in \mathbb{R}^{|\mathcal{R}_\Psi|}\) is the learned weight vector, and \(b \in \mathbb{R}\) is the bias term and \(g(z) = \frac{1}{1 + \exp(-z)}\).
 \(w\) and \(b\) are obtained by minimizing the following regularized logistic loss:
 \vspace{-5pt}
{\small
\[
\min_{w, b} \; \frac{1}{\vert L^{\prime} \vert} \sum_{i=1}^{\vert L^{\prime} \vert} \log\left(1 + \exp\left(-y_i \left(w^\top \phi(\sigma_i) + b\right)\right)\right) + \lambda \|w\|_1,
\]}
where {\small$\|w\|_1 {=} \sum_{j=1}^{\vert \mathcal{R}_{\Psi}\vert} |w_j|$}, {\small\(\sigma_i {\in} L^{\prime}\)} are training traces, {\small \(y_i {=} \kappa_L(\sigma_i)\)} are the corresponding labels, and {\small\(\lambda > 0\)} is the regularization strength controlling the sparsity of \(w\).

We consider the weights learned by the logistic regression model as the importance of each rule $R_j \in \mathcal{R}_{\Psi}$, denoted as $importance(R_j) = w[j]$ where $1 \leq j \leq \vert \mathcal{R}_{\Psi} \vert$.
\end{definition}

\subsection{Hierarchical Clustering and Rule Selection}
Although the important rules extracted from the regression model may overlap in the traces they cover, we aim to identify a representative and diverse subset of them. To this end, we apply hierarchical clustering to group rules that refer to similar sets of traces. As a dissimilarity measure, we use the \emph{Jaccard distance} between binary feature vectors, allowing us to cluster and select distinctive rules more effectively.

\begin{definition}[Jaccard Distance between Rules]
Let $L \in \mathcal{L}$ be an event log and $R_i, R_j \in \mathcal{R}_{\Psi}$ be two rules extracted from the ensemble of trees where $1\leq i,j \leq \vert \mathcal{R}_{\Psi} \vert$. We define the Jaccard distance between rules $R_i$ and $R_j$ as:
{\small
\[
Jac(R_i, R_j) = 1-\frac{\underset{\sigma \in L}{\sum}\big(\phi(\sigma)[i] \wedge \phi(\sigma)[j]\big)}{\underset{\sigma \in L}{\sum}\big(\phi(\sigma)[i] \vee \phi(\sigma)[j]\big)},
\]}
\end{definition}

After computing the pairwise Jaccard distances between all rule pairs, we apply agglomerative hierarchical clustering, iteratively merging clusters of rules that exhibit the minimal average inter-cluster distance
    {\small
    \[
    dist(G, G^{\prime}) {=} \frac{1}{|G| \cdot |G^{\prime}|} \sum_{R \in G} \sum_{R^{\prime} \in G^{\prime}} Jac(R, R^{\prime}),
    \]
    } where $G$ and $G^{\prime}$ are two clusters of rules.

Once clusters of similar rules are obtained, we proceed by selecting a representative rule for each cluster. Specifically, for each cluster $G {\subseteq} \mathcal{R}_{\Psi}$, we choose the rule with the highest importance as the representative rule:
{\small \[
R^{*}_G {=} \text{argmax}_{R_j \in G} |importance(R_j)|,
\]}
where {\small $1 {\leq} j {\leq} \vert \mathcal{R}_{\Psi} \vert$}. Subsequently, we filter the original set of traces to retain only those that satisfy the chosen representative rule $R^{*}_G$, yielding the subset of traces
{\small \(
L|_G {=} \{\sigma \in L \mid \phi(\sigma)[j] {=} 1 \text{ where } R_j {=} R^{*}_G\}.
\)}

\section{Evaluation Metrics}
Given a set of \( K\) rule clusters, we focus on each cluster \( G_k \) where \( 1 \leq k \leq K \) and the corresponding filtered event log \( L|_{G_k} \), which contains traces satisfying the representative discriminative rule of that cluster. For each \( L|_{G_k} \), we independently apply process discovery to derive a process model that captures the behavior specific to the associated traces. This results in interpretable, cluster-specific models that not only describe the observed behavior but also facilitate distinguishing between desirable and undesirable cases.

To assess the representativeness of the models, we employ well-established conformance checking techniques~\cite{DBLP:books/sp/CarmonaDSW18} including
 \textit{trace fitness} ($t\mhyphen fit$), i.e., the percentage of traces that perfectly fit the model.
 \textit{Alignment fitness} ($a\mhyphen fit$), i.e., the weighted average of alignment-based fitness values across traces.
\textit{Precision} ($prc$), i.e., an escaping-edges-based measure that quantifies the amount of behavior allowed by the model but not observed in the event log.

Furthermore, we evaluate the discriminative power of the models using metrics introduced in~\cite{DBLP:conf/sac/NorouzifarA23}, which are defined in terms of trace and alignment fitness. 

\begin{definition}[Discriminative Evaluation Metrics]
Let \( L \in \mathcal{L} \) be an event log and \( \kappa_L \) a labeling function that partitions \( L \) into desirable and undesirable traces:
{\small
\(
L^+ {=} \{\sigma {\in} L \mid \kappa_L(\sigma) {=} 1\} 
\)}
and
{\small
\(
L^- {=} \{\sigma {\in} L \mid \kappa_L(\sigma) {=} 0\}.
\)}
Let \(\mathcal{M} \) be the universe of Petri net models and \( M {\in} \mathcal{M} \) denote a Petri net model. We define the complement of the fitness measures as
{\small
\(
\overline{a\text{-}fit}(L, M) = 1 - a\text{-}fit(L, M) \quad \text{and} \quad \overline{t\text{-}fit}(L, M) = 1 - t\text{-}fit(L, M).
\)}

Alignment-based accuracy ($a\text{-}acc$), trace-based accuracy ($t\text{-}acc$), alignment-based F1-score ($a\text{-}F1$), and trace-based F1-score ($t\text{-}F1$)  are defined as follows:
\begin{itemize}
    \item 
    \(
    a\text{-}acc(L^+, L^-, M) = a\text{-}fit(L^+, M) - a\text{-}fit(L^-, M)\)
    \item
    \(
    t\text{-}acc(L^+, L^-, M) = t\text{-}fit(L^+, M) - t\text{-}fit(L^-, M)\)
    \item
    \(
    a\text{-}F1(L^+, L^-, M) = \frac{2 \times a\text{-}fit(L^+, M) \times \overline{a\text{-}fit}(L^-, M)}{a\text{-}fit(L^+, M) + \overline{a\text{-}fit}(L^-, M)}
    \)
    \item
    \(
    t\text{-}F1(L^+, L^-, M) = \frac{2 \times t\text{-}fit(L^+, M) \times \overline{t\text{-}fit}(L^-, M)}{t\text{-}fit(L^+, M) + \overline{t\text{-}fit}(L^-, M)}
    \)
\end{itemize}
\end{definition}

The metrics {\small $a\text{-}acc(L^+, L^-, M)$} and {\small $t\text{-}acc(L^+, L^-, M)$} range from $-1$ to $1$. A value closer to $1$ indicates that the alignment or trace fitness is substantially higher with respect to the desirable event log.
The metrics {\small $a\text{-}F1(L^+, L^-, M)$} and {\small $t\text{-}F1(L^+, L^-, M)$} range from $0$ to $1$, where a value of $1$ denotes that the model balances well between representing desirable event log and avoiding the undesirable event log.
The accuracy metrics primarily emphasize the difference in fitness values between the groups, whereas the F1 measures focus on the trade-off between increasing fitness for the desirable log and reducing it for the undesirable log. 
For example, if {\small\( a\text{-}fit(L^+, M) {=} 1 \)} and {\small\( a\text{-}fit(L^-, M) {=} 1 \)}, then {\small\( a\text{-}acc(L^+, L^-, M) {=} 0 \)}, indicating a mid-range value. However, {\small\( a\text{-}F1(L^+, L^-, M) {=} 0 \)}, which is the minimum value, as the perfect fitness with respect to \( L^- \) means that one of the main objectives that is avoiding the undesirable behavior is sacrificed.

\vspace{-15 pt}
\section{Evaluation}
The proposed framework has been fully implemented and is publicly available\footnote{\url{https://github.com/aliNorouzifar/discriminative_process_discovery}}. We evaluated the framework using several real-life event logs. This section presents an overview of the event logs used, the experimental setup, key findings, and a discussion of the framework’s limitations.

\vspace{-5 pt}
\subsection{Event logs}
 Since the choice of \(L^+\) and \(L^-\) depends on the specific application context, we adopted the method proposed in~\cite{DBLP:conf/bpmds/NorouzifarRDA24} to construct these labeled logs for process discovery. This method ranks traces based on a selected performance indicator and identifies significant control-flow changes using a two-sided sliding window in combination with the earth mover’s distance. We selected case duration as the performance dimension and used the identified change points to define the desirable and undesirable event logs. The resulting trace groups are summarized in Table~\ref{table:event_logs}.

\begin{table}[tb]
\caption{\small The event logs used in the experiments.}
\label{table:event_logs}
\centering
\begin{tabular}{|c|l|c|}
\hline
\textbf{Event Log} & \multicolumn{1}{c|}{\textbf{Desirable/Undesirable Event Logs}} & \textbf{N. Traces} \\ \hline
\multirow{2}{*}{BPIC12}           
    & $L^+$: duration 4 to 28 days         & 3719  \\ \cline{2-3}
    & $L^-$: duration more than 28 days    & 1433  \\ \hline
\multirow{2}{*}{BPIC17}           
    & $L^+$: duration more than 28 days    & 20282 \\ \cline{2-3}
    & $L^-$: duration less than 28 days    & 11227 \\ \hline
\multirow{2}{*}{Hospital Billing} 
    & $L^+$: duration more than 1 day      & 74806 \\ \cline{2-3}
    & $L^-$: duration less than 1 day      & 25194 \\ \hline
\end{tabular}
\vspace{-10pt}
\end{table}

\vspace{-5 pt}
\subsection{Experimental Setup}
For each event log, we used 70\% of the traces for feature extraction, training the ensemble models and logistic regression classifier, and reserved the remaining 30\% for testing.
    Since class imbalance can negatively impact the learning process, we applied undersampling in training to ensure equal representation of desirable and undesirable cases.
     We evaluated different parameter settings for the supervised learning models using 5-fold cross-validation on the training data to determine the optimal configuration. It includes different random forest and gradient boosting settings for the rule generation and regression model parameters.

We applied agglomerative hierarchical clustering and determined the appropriate number of clusters by visually inspecting the resulting dendrograms.
 To avoid relying on the representational bias of a single discovery technique, we experimented with multiple process discovery algorithms:
    \begin{itemize}
        \item Rule-guided Inductive Miner (IMr) with a support threshold of \( \mathit{sup} = 0.2 \)~\cite{DBLP:conf/rcis/NorouzifarDA24}. The parameter settings for the discovery of declarative rules is set to $support = 0.005$ and $confidence = 0.95$. 
        \item Inductive Miner with infrequency filtering (IMf) with a frequency threshold of \( f = 0.2 \)~\cite{DBLP:conf/bpm/LeemansFA13}.
        \item Split Miner (SM) with parameters \( \eta {=} 0.4 \) and \( \epsilon {=} 0.1 \)~\cite{DBLP:journals/kais/AugustoCDRP19}.
    \end{itemize}

 \begin{figure}[htb]
    \centering
    \begin{subfigure}{0.47\linewidth}
    \centering
    \includegraphics[width=1\linewidth]{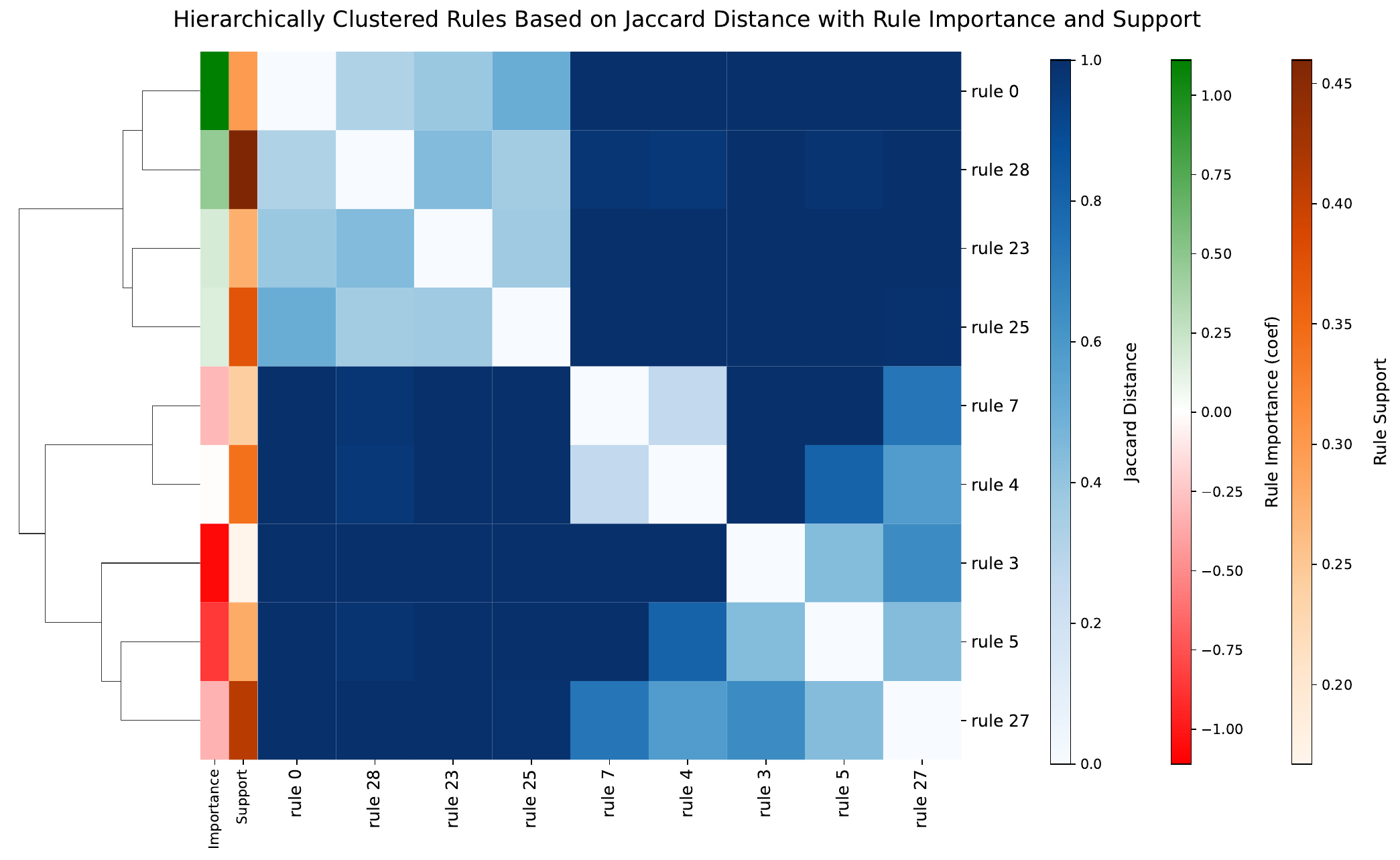}
    \caption{\small BPIC 2012 log}
    \label{fig:heatmap_BPIC12}
\end{subfigure}
\begin{subfigure}{0.47\linewidth}
    \centering
    \includegraphics[width=1\linewidth]{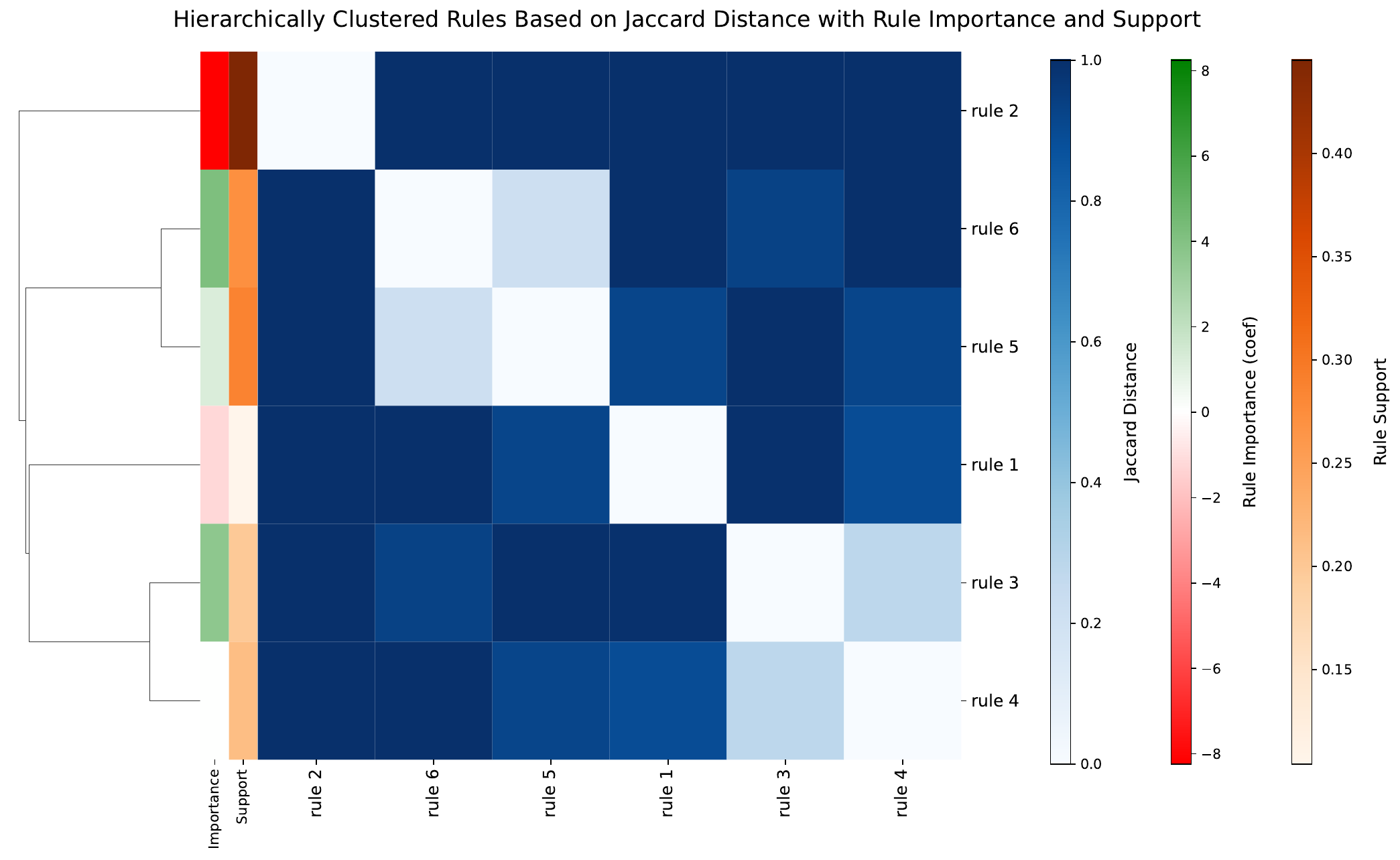}
    \caption{\small Hospital Billing log}
    \label{fig:heatmap_HB}
\end{subfigure}\\
\begin{subfigure}{0.5\linewidth}
    \centering
    \includegraphics[width=1\linewidth]{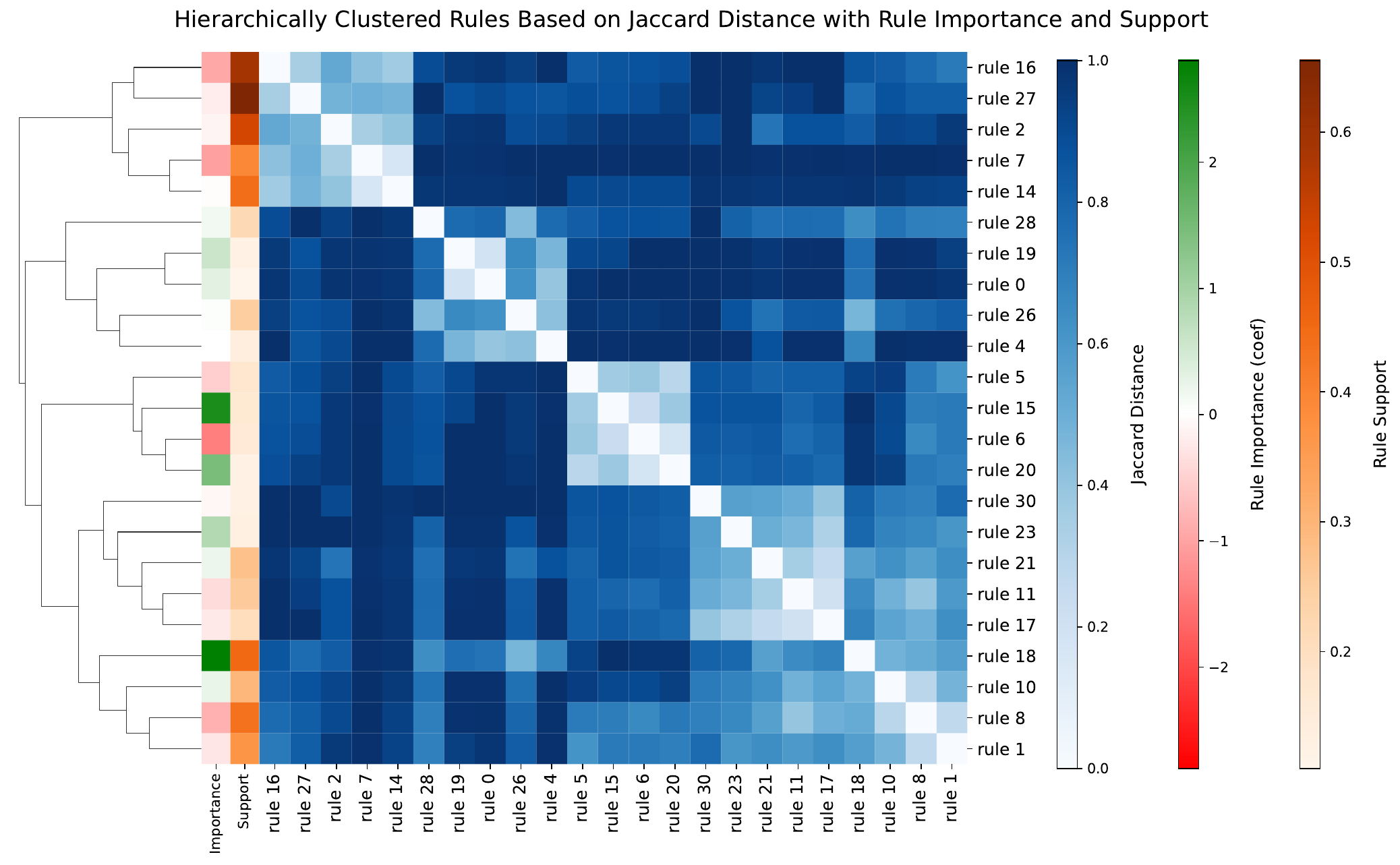}
    \caption{\small BPIC 2017 log}
    \label{fig:heatmap_BPIC17}
\end{subfigure}
\caption{\small 
Jaccard distance heatmaps between extracted rules for each event log. Lighter blue indicates greater similarity. Rule support (orange) and importance (green-red) are shown on the left. Dendrograms guide the identification of trace clusters.
}
\label{fig:heatmap}
\vspace{-15pt}
\end{figure}

\vspace{-20pt}
\subsection{Analysis of the Results}

Figure~\ref{fig:heatmap} represents heatmaps of the Jaccard distances between rules extracted for each event log. The rows and columns correspond to rules obtained from regression models after initial feature generation using declarative constraints and ensemble tree-based rule extraction. The intensity of the blue color reflects the magnitude of the Jaccard distance, i.e., \(Jac(R,R^{\prime}) \) where $R$ is the rule corresponding to the row and $R^{\prime}$ is the rule correponding to the column, so that lighter shades indicate that the corresponding rules are satisfied by similar groups of traces. In addition to the distance information, the heatmap includes a column on the left displaying the support for each rule in orange, and another column representing the importance magnitude using a color scale ranging from green to red with darker green denoting high positive coefficients and darker red indicating high negative coefficients. The dendrogram on the left, derived from hierarchical clustering based on the Jaccard distance, helps to determine an appropriate number of clusters for achieving well-separable groups of traces.

Based on the dendrograms represented in Fig.~\ref{fig:heatmap}, we choose the number of clusters 3 for the BPIC2012, 4 for the hospital billing, and 4 for the BPIC2017 log. 
In Table~\ref{table:summary}, we report the accuracy of the trained machine learning models for each event log (\({ML\text{-}acc}\)), that is the ratio of the correctly classified traces (true positives and true negatives) to the number of traces.  For every extracted cluster, the most important rule is shown in the table, along with its alignment accuracy, trace accuracy, importance magnitude, and support values in both the desirable and undesirable event logs. The reported alignment and trace accuracy values are calculated considering the models discovered using the IMr discovery technique.

Figure~\ref{fig:evaluation_metrics} presents a comparative overview of various evaluation metrics computed for different trace groups using models discovered by three process discovery techniques. Specifically, Figures~\ref{fig:evaluation_metrics_IMr},~\ref{fig:evaluation_metrics_IMf}, and~\ref{fig:evaluation_metrics_SM} illustrate the results for models obtained using the IMr, IMf, and SM techniques, respectively. Each subfigure visualizes a distinct evaluation metric calculated across the three event logs used in the study. Trace groups, represented by different colors, include both the clusters identified in Fig.~\ref{fig:heatmap} and described in Table~\ref{table:summary}, as well as the baseline groups composed solely of desirable traces and undesirable traces. 

\subsubsection{BPIC 2012}
\begin{figure}[H]
    \centering
        \begin{subfigure}{0.32\linewidth}
    \centering
        \includegraphics[width=1\linewidth]{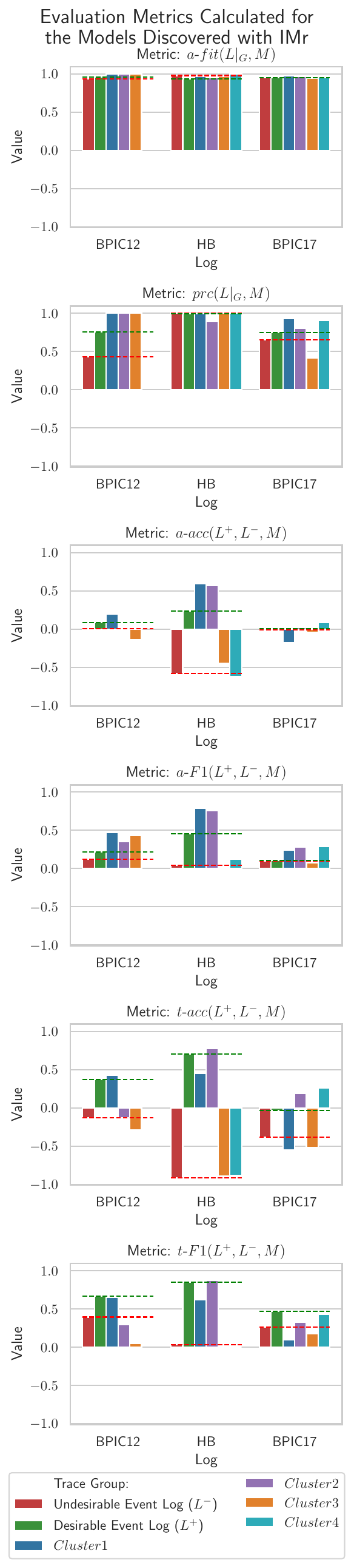}
         \caption{\small Models discovered with the IMr process discovery technique.}
    \label{fig:evaluation_metrics_IMr}
    \end{subfigure}
    \begin{subfigure}{0.32\linewidth}
    \centering
    \includegraphics[width=1\linewidth]{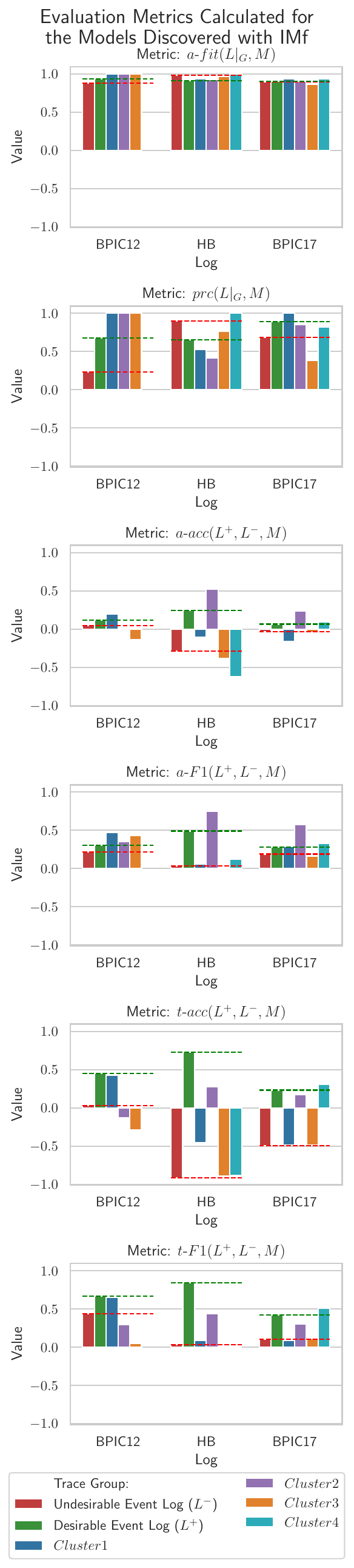}
    \caption{\small Models discovered with the IMf process discovery technique.}
    \label{fig:evaluation_metrics_IMf}
\end{subfigure}
\begin{subfigure}{0.32\linewidth}
    \centering
    \includegraphics[width=1\linewidth]{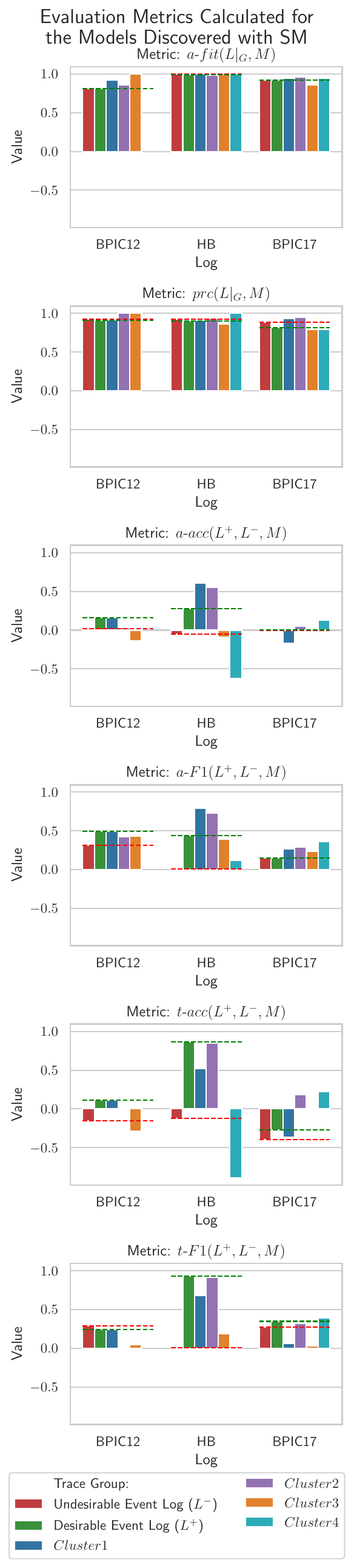}
    \caption{\small Models discovered with the SM process discovery technique.}
    \label{fig:evaluation_metrics_SM}
\end{subfigure}
    \caption{\small Evaluation metrics for discovered models across trace groups. Bar heights show values based on IMr (left), IMf (middle), and Split Miner (right). Results are grouped by evaluation metric for each cluster and baseline (\(L^+\) and \(L^-\)).}
    \label{fig:evaluation_metrics}
\end{figure}

\begin{figure}[htb]
    \centering
    \begin{subfigure}{1\linewidth}
    \centering
    \includegraphics[width=0.9\linewidth]{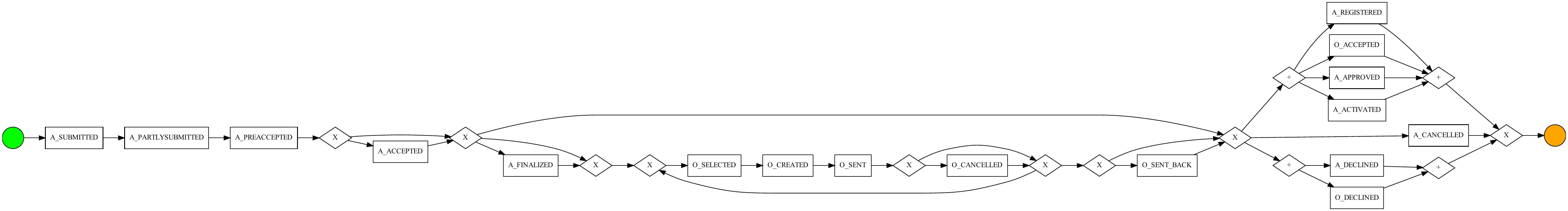}
    \caption{\small Discovered model from the undesirable event log}
    \label{fig:BPIC12_undesirable_model}
\end{subfigure}\\
\begin{subfigure}{1\linewidth}
    \centering
    \includegraphics[width=0.9\linewidth]{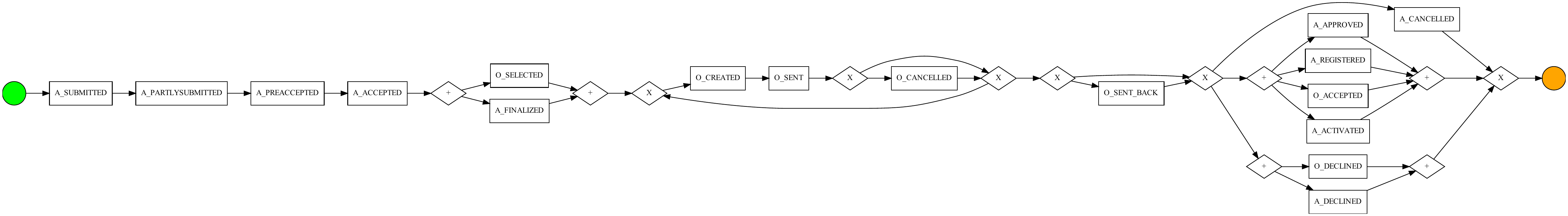}
    \caption{\small Discovered model from the desirable event log}
    \label{fig:BPIC12_desirable_model}
\end{subfigure}\\
    \begin{subfigure}{1\linewidth}
    \centering
        \includegraphics[width=0.7\linewidth]{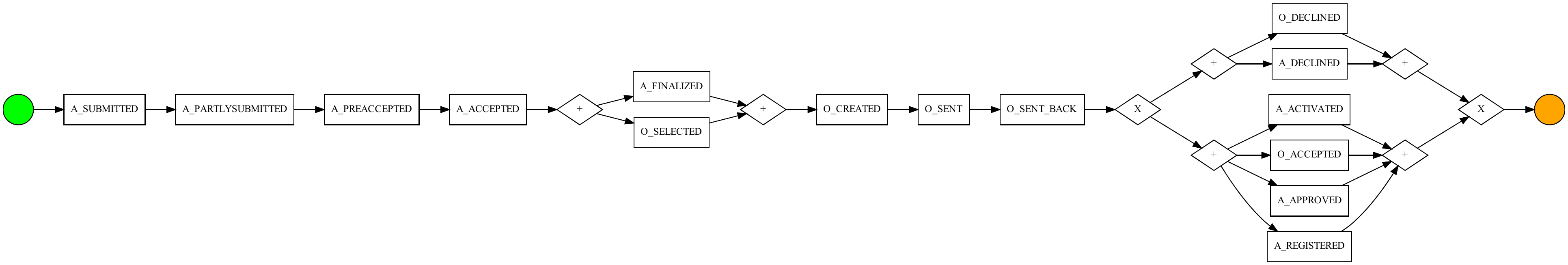}
         \caption{\small Discovered model from cluster 1}
    \label{fig:BPIC12_cluster1_model}
    \end{subfigure}\\
    \begin{subfigure}{0.65\linewidth}
    \centering
        \includegraphics[width=1\linewidth]{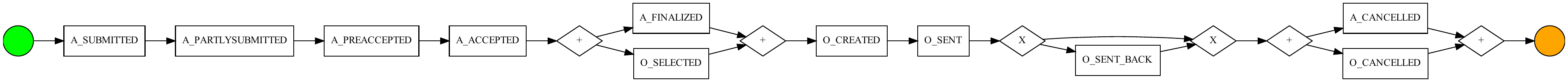}
         \caption{\small Discovered model from cluster 2}
    \label{fig:BPIC12_cluster2_model}
    \end{subfigure}
    \begin{subfigure}{0.3\linewidth}
    \centering
        \includegraphics[width=1\linewidth]{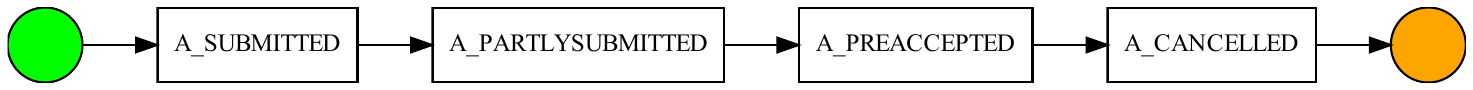}
         \caption{\small Discovered model from cluster 3}
    \label{fig:BPIC12_cluster3_model}
    \end{subfigure}
    \caption{\small The discovered model from BPIC 12 event log using the IMr process discovery algorithm.}
    \label{fig:BPIC12_models}
    \vspace{-15pt}
\end{figure}

As shown in Fig.~\ref{fig:evaluation_metrics}, the models discovered considering the representative rule of cluster 1 achieve higher $a\mhyphen acc$ and $a\mhyphen F1$ scores compared to the baseline model derived from the desirable event log in both the IMr and IMf experiments. The $t\mhyphen acc$ and $t\mhyphen F1$ scores for cluster 1 may be slightly higher or lower, depending on the experiment. Considering Table~\ref{table:summary}, the most important rule in cluster 1 indicates that in the corresponding traces O\_Created is not followed by O\_Selected and A\_Activated and A\_Declined do not coexist. The illustrated model in Fig.~\ref{fig:BPIC12_cluster1_model} shows the discovered model based on this event log that contains the control flow corresponding to the accepted and denied cases.

\begin{table}[H]
\vspace{-20pt}
\centering
\caption{\small Most important rule per discovered cluster with alignment accuracy, trace accuracy, coefficient, and support in $L^+$/$L^-$. Accuracy corresponds to the classifier trained for each event log; discovery and evaluation use the IMr technique.}
\label{table:summary}
\begin{tabular}{|l|c|l|lcccl|}
\hline
\multirow{2}{*}{\textbf{Log}} & \multirow{2}{*}{\(\mathbf{ML\text{-}acc}\)} & \multirow{2}{*}{\textbf{Cluster}} & \multicolumn{5}{l|}{\textbf{Rule}}                                                                                                                                                                                                          \\ \cline{4-8} 
                              &                                             &                                   & \multicolumn{1}{p{1.5cm}|}{\(\mathbf{a\text{-}acc}\)} & \multicolumn{1}{p{1.5cm}|}{\(\mathbf{t\text{-}acc}\)} & \multicolumn{1}{p{1.5cm}|}{\(\mathbf{Coef}\)} & \multicolumn{1}{p{1.7cm}|}{\(\mathbf{sup}(L^+)\)} & \multicolumn{1}{p{1.7cm}|}{\(\mathbf{sup}( L^-)\)} \\ \hline
\multirow{6}{*}{BPIC12}       & \multirow{6}{*}{0.74}                       & \multirow{2}{*}{1 (rule 0)}       & \multicolumn{5}{l|}{\begin{tabular}[c]{@{}l@{}}$(NotSuccession(\text{O\_Cre},\text{O\_Sel}),satisfied) {=}1 \wedge$\\ $(NotCoExistence(\text{A\_Act},\text{A\_Dec}),satisfied) {=}1$\end{tabular}}                                                                 \\ \cline{4-8} 
                              &                                             &                                   & \multicolumn{1}{c|}{0.19}                      & \multicolumn{1}{c|}{0.43}                      & \multicolumn{1}{c|}{1.11}              & \multicolumn{1}{c|}{0.50}                      & 0.07                                            \\ \cline{3-8} 
                              &                                             & \multirow{2}{*}{2 (rule 7)}       & \multicolumn{5}{l|}{\begin{tabular}[c]{@{}l@{}}($AlternateSuccession(\text{A\_Fin},\text{O\_Can}),satisfied) {=}1 \wedge$ \\$(NotCoExistence(\text{A\_Pre},\text{A\_Can}),violated) {=}1$\end{tabular}}                                                                \\ \cline{4-8} 
                              &                                             &                                   & \multicolumn{1}{c|}{0.00}                      & \multicolumn{1}{c|}{-0.12}                     & \multicolumn{1}{c|}{-0.31}             & \multicolumn{1}{c|}{0.19}                      & 0.32                                            \\ \cline{3-8} 
                              &                                             & \multirow{2}{*}{3 (rule 3)}       & \multicolumn{5}{l|}{\begin{tabular}[c]{@{}l@{}}$(AlternateSuccession(\text{A\_Sub},\text{O\_SentB}),satisfied) {=}0 \wedge$ \\ $(AlternateSuccession(\text{A\_Acc},\text{A\_Dec}),vac\mhyphen satisfied) {=}1$\end{tabular}}                                                              \\ \cline{4-8} 
                              &                                             &                                   & \multicolumn{1}{c|}{-0.14}                     & \multicolumn{1}{c|}{-0.29}                     & \multicolumn{1}{c|}{-1.07}             & \multicolumn{1}{c|}{0.02}                      & 0.31                                            \\ \hline
\multirow{8}{*}{HB}           & \multirow{8}{*}{0.95}                       & \multirow{2}{*}{1 (rule 6)}       & \multicolumn{5}{l|}{\begin{tabular}[c]{@{}l@{}}$(NotCoExistence(\text{CHD},\text{COO}),violated) {=}1 \wedge$ \\ $(AlternateSuccession(\text{FIN},\text{JP}),violated) {=}1$\end{tabular}}                                                                             \\ \cline{4-8} 
                              &                                             &                                   & \multicolumn{1}{c|}{0.59}                      & \multicolumn{1}{c|}{0.45}                      & \multicolumn{1}{c|}{4.13}              & \multicolumn{1}{c|}{0.54}                      & 0.00                                            \\ \cline{3-8} 
                              &                                             & \multirow{2}{*}{2 (rule 3)}       & \multicolumn{5}{l|}{\begin{tabular}[c]{@{}l@{}}$(NotCoExistence(\text{FIN},\text{JP}),vac\mhyphen satisfied) {=}0 \wedge$ \\ $(AlternateSuccession(\text{CHD},\text{BILLED}),violated) {=}1$\end{tabular}}                                                                         \\ \cline{4-8} 
                              &                                             &                                   & \multicolumn{1}{c|}{0.57}                      & \multicolumn{1}{c|}{0.77}                      & \multicolumn{1}{c|}{3.64}              & \multicolumn{1}{c|}{0.40}                      & 0.00                                            \\ \cline{3-8} 
                              &                                             & \multirow{2}{*}{3 (rule 1)}       & \multicolumn{5}{l|}{\begin{tabular}[c]{@{}l@{}}$(ChainSuccession(\text{RLS},\text{STAT}),violated) {=}0 \wedge$ \\ $(End(\text{NEW}),violated) {=}1$\end{tabular}}                                                                                         \\ \cline{4-8} 
                              &                                             &                                   & \multicolumn{1}{c|}{-0.45}                     & \multicolumn{1}{c|}{-0.89}                     & \multicolumn{1}{c|}{-1.26}             & \multicolumn{1}{c|}{0.11}                      & 0.11                                            \\ \cline{3-8} 
                              &                                             & \multirow{2}{*}{4 (rule 2)}       & \multicolumn{5}{l|}{$(End(\text{NEW}),satisfied) {=}1$}                                                                                                                                                                                            \\ \cline{4-8} 
                              &                                             &                                   & \multicolumn{1}{c|}{-0.62}                     & \multicolumn{1}{c|}{-0.89}                     & \multicolumn{1}{c|}{-8.24}             & \multicolumn{1}{c|}{0.00}                      & 0.89                                            \\ \hline
\multirow{8}{*}{BPIC17}       & \multirow{8}{*}{0.83}                       & \multirow{2}{*}{1 (rule 7)}       & \multicolumn{5}{l|}{$(ChainSuccession(\text{A\_Val},\text{A\_Den}),vac\mhyphen satisfied) {=}1$}                                                                                                                                                                    \\ \cline{4-8} 
                              &                                             &                                   & \multicolumn{1}{c|}{-0.17}                     & \multicolumn{1}{c|}{-0.55}                     & \multicolumn{1}{c|}{-1.05}             & \multicolumn{1}{c|}{0.07}                      & 0.71                                            \\ \cline{3-8} 
                              &                                             & \multirow{2}{*}{2 (rule 19)}      & \multicolumn{5}{l|}{\begin{tabular}[c]{@{}l@{}}$(CoExistence(\text{A\_Can},\text{O\_Sent}),satisfied) {=}0 \wedge$ \\ $(ChainSuccession(\text{O\_Ret},\text{O\_Acc}),violated) {=}0$\end{tabular}}                                                                \\ \cline{4-8} 
                              &                                             &                                   & \multicolumn{1}{c|}{0.03}                      & \multicolumn{1}{c|}{0.19}                      & \multicolumn{1}{c|}{0.58}              & \multicolumn{1}{c|}{0.24}                      & 0.02                                            \\ \cline{3-8} 
                              &                                             & \multirow{2}{*}{3 (rule 15)}      & \multicolumn{5}{l|}{\begin{tabular}[c]{@{}l@{}}$(CoExistence(\text{A\_Val},\text{A\_Comp}),violated) {=}0 \wedge$ \\ $(End(\text{O\_Can}),satisfied) {=}1$\end{tabular}}                                                                                    \\ \cline{4-8} 
                              &                                             &                                   & \multicolumn{1}{c|}{-0.04}                     & \multicolumn{1}{c|}{-0.52}                     & \multicolumn{1}{c|}{2.51}              & \multicolumn{1}{c|}{0.19}                      & 0.14                                            \\ \cline{3-8} 
                              &                                             & \multirow{2}{*}{4 (rule 18)}      & \multicolumn{5}{l|}{$(End(\text{O\_Can}),violated) {=}1$}                                                                                                                                                                                         \\ \cline{4-8} 
                              &                                             &                                   & \multicolumn{1}{c|}{0.09}                      & \multicolumn{1}{c|}{0.26}                      & \multicolumn{1}{c|}{2.80}              & \multicolumn{1}{c|}{0.75}                      & 0.15                                            \\ \hline
\end{tabular}
\vspace{-20pt}
\end{table}

For the undesirable traces, models discovered from cluster~3 consistently outperform the baseline across all experiments by achieving lower $a\mhyphen acc$, $t\mhyphen acc$, and $t\mhyphen F1$ values. The increase in the \( a\text{-}F1 \) score is due to a shift from a situation where {\small \( a\text{-}fit(L^-, M) \)} is disproportionately high to a situation where a better balance between {\small \( a\text{-}fit(L^-, M) \)} and {\small \( a\text{-}fit(L^+, M) \)} is achieved. For example, in the IMr experiment, {\small \( a\text{-}fit(L^+, M) \)} decreases from 0.94 to 0.39, and {\small \( a\text{-}fit(L^-, M) \)} from 0.93 to 0.53. While \( a\text{-}acc \) decreased, indicating greater distinguishability, the score \( a\text{-}F1 \) increases due to the more balanced representation of desirable and avoidance of undesirable traces.

Cluster 3 includes the traces in which the alternative succession of A\_Accepted and A\_Declined is vacuously satisfied. The model illustrated in Fig.~\ref{fig:BPIC12_cluster3_model} shows the discovered model for this event log that contains the control flow pertain to the canceled applications. Similarly, cluster 2 with the discovered model represented in Fig.~\ref{fig:BPIC12_cluster2_model} shows the control flow for a bigger group of canceled applications. Comparing these models to the discovered models based on the desirable event log, i.e., Fig.~\ref{fig:BPIC12_desirable_model} and the undesirable event log, i.e., Fig.~\ref{fig:BPIC12_undesirable_model} show that the discovered models from the clusters more specifically represents the differences between the desirable and undesirable cases.

\subsubsection{Hospital Billing}
The evaluation metrics presented in Fig.~\ref{fig:evaluation_metrics} highlight that, the models discovered based on the representative rule of cluster 2 outperform those derived from the desirable event log in terms of all four metrics $a\mhyphen acc$, $a\mhyphen F1$, $t\mhyphen acc$, and $t\mhyphen F1$ across the IMr and SM experiments. Similarly, models discovered using the representative rule of cluster 1 yield higher $a\mhyphen acc$ and $a\mhyphen F1$ scores compared to the baseline. However, this improvement comes at the cost of slightly reduced $t\mhyphen acc$ and $t\mhyphen F1$ scores.

 Although different process discovery techniques use the same group of traces to discover process models, the quality of the discovered models and the extent to which they generalize the observed behavior can vary significantly due to the representational bias of each discovery technique. In cluster 1, models discovered using IMr and SM demonstrate substantially higher alignment accuracy and F1-score compared to those discovered solely from the desirable event log. However, models generated using IMf tend to overgeneralize the behavior, allowing both desirable and undesirable traces to fit the model. This leads to lower scores in evaluation metrics that assess discriminative power. The low precision of IMf models further suggests that these models permit behaviors not present in the event logs. While this may indicate flexibility, it does not clarify whether the additional allowed behavior reflects acceptable generalization or undesirable drift.

Clusters 4 and 3 correspond to behaviors that are more representative of the undesirable event log. However, their performance does not consistently surpass that of the models discovered directly from the undesirable event log. The evaluation metrics across different experiments indicate that, in some cases, the models based on these clusters achieve better scores than the baseline, but this superiority is not observed uniformly across all settings.

\subsubsection{BPIC 2017}
Based on the evaluation metrics shown in Fig.~\ref{fig:evaluation_metrics}, the models discovered from different trace groups reveal interesting patterns. Clusters 1 and 3 are particularly effective in capturing undesirable behavior, while clusters 2 and 4 are better suited for representing desirable behavior. For example, in the IMr experiments, the model derived from cluster 1 exhibits significantly lower $a\mhyphen acc$, $t\mhyphen acc$, and $t\mhyphen F1$ scores compared to the undesirable log, indicating a strong alignment with undesirable traces. Cluster 3 follows a similar pattern, though with less significant differences. On the other hand, the model from cluster 4 achieves significantly higher $a\mhyphen acc$, $a\mhyphen F1$, and $t\mhyphen acc$ scores compared to the model discovered from the desirable event log. Cluster 2 shows a comparable trend, though the improvements are less substantial.

Depending on the interdependencies among features and their discriminative strength, the number of important features identified by the sparse regression model trained on desirable and undesirable event logs may vary. Features identified as important and clustered together often exhibit consistent directional influence, generally highlighting either desirable or undesirable behavior. However, this consistency does not always hold. For instance, in the BPIC 2017 event log, we observe that rule 15 and rule 6, although referring to similar groups of traces, receive coefficients with opposite signs. This discrepancy may stem from subtle behavioral differences that are particularly relevant for the classification task. Alternatively, it may be a consequence of multicollinearity among features, which warrants further investigation.

A concrete example of this can be seen in cluster 3 of Fig.~\ref{fig:heatmap_BPIC17} and Table~\ref{table:summary}. While the most important feature in this cluster has a relatively high positive coefficient, a very similar feature in the heatmap exhibits a strong negative coefficient. This contrast suggests potential multicollinearity that could destabilize the interpretation of individual coefficients. Notably, the conformance checking results for models discovered using this cluster across different discovery techniques indicate that the resulting models better capture undesirable behavior, despite the most important rule suggesting desirability.

This observation underscores the need for caution when interpreting feature importance in isolation. We therefore recommend focusing primarily on clusters where the important features have a consistent directional interpretation, such as cluster 1 and cluster 2, which offer clearer insights into the behavioral differences between trace groups. In contrast, cluster 4 includes a rule with a strong positive coefficient that does not align closely with others in the heatmap. However, conformance checking suggests that this rule still corresponds well to traces exhibiting desirable behavior, indicating that the differences between the traces corresponding to this feature and other features within this group is contributing to the discriminative power of the models.

Based on the experimental results across scenarios with both desirable and undesirable event logs, discovering process models that consider only a single aspect is less effective in explaining the differences between the logs. In contrast, the framework proposed in this paper facilitates the identification of the aspects that drive these distinctions. The process models subsequently discovered for these aspects are better suited to represent the distinctive behaviors within each event log.

\section{Conclusion}
The proposed method leverages the strengths of supervised learning to identify key differences between desirable and undesirable event logs. These discriminative features enable the discovery of process models that are more focused on representing the underlying factors that drive this distinction. The results are promising and introduce a novel perspective for working with labeled event logs, supporting interpretable and outcome-aware process discovery. 

Nonetheless, the approach also presents opportunities for further refinement. In particular, multicollinearity among features may lead to instability in the regression coefficients, an issue that warrants further investigation and can be mitigated through preprocessing techniques. Using discriminative patterns to first filter the event log and then discover imperative models, which are subsequently evaluated for their relevance to desirable and undesirable traces, represents an indirect approach to identifying discriminative process models. Future work could explore more direct methods for deriving such subprocess models.

%
%
%
\bibliographystyle{splncs04}
\bibliography{lit}

\end{document}